\pdfoutput=1

\documentclass[11pt]{article}

\usepackage[final]{acl}

\usepackage{times}
\usepackage{latexsym}

\usepackage[T1]{fontenc}
\usepackage{tcolorbox}
\usepackage{multirow}

\usepackage[utf8]{inputenc}
\usepackage{subcaption}  

\usepackage{microtype}

\usepackage{inconsolata}

\usepackage{graphicx}
\usepackage{forest}
\usepackage{enumitem}
\newtcolorbox{llmprompt}[1][]{
  colback=gray!20, 
  colframe=gray!70,
  fonttitle=\bfseries\ttfamily,
  title=#1, 
  rounded corners, 
  fontupper=\tiny\ttfamily,
  before upper={\raggedright}
}

\title{SELF-PERCEPT: Introspection Improves Large Language Models’ Detection of Multi-Person Mental Manipulation in Conversations}

\author{Danush Khanna$^{1}$ \qquad Pratinav Seth$^{2,3}$ \qquad Sidhaarth Murali$^{4}$ \\ 
\textbf{Aditya Guru}$^{1}$ \qquad \textbf{Siddharth Shukla}$^{1}$ \qquad \textbf{Tanuj Tyagi}$^{1}$ \\ 
\textbf{Sandeep Chaurasia}$^{1}$ \qquad \textbf{Kripabandhu Ghosh}$^{5}$ \\ 
$^{1}$Manipal University Jaipur, India \quad
$^{2}$Manipal Institute of Technology, India \quad \\
$^{3}$AryaXAI Alignment Lab, AryaXAI.com, India \\
$^{4}$National Institute of Technology Karnataka, Surathkal, India \quad
$^{5}$IISER Kolkata, India \\ 
\texttt{danush.229310455@muj.manipal.edu}
}

\begin{document}

\maketitle
\begin{abstract}
Mental manipulation is a subtle yet pervasive form of abuse in interpersonal communication, making its detection critical for safeguarding potential victims. However, due to manipulation's nuanced and context-specific nature, identifying manipulative language in complex, multi-turn, and multi-person conversations remains a significant challenge for large language models (LLMs).
To address this gap, we introduce the MultiManip dataset, comprising 220 multi-turn, multi-person dialogues balanced between manipulative and non-manipulative interactions, all drawn from reality shows that mimic real-world scenarios. 
For manipulative interactions, it includes 11 distinct manipulations depicting real-life scenarios.
We conduct extensive evaluations of state-of-the-art LLMs, such as GPT-4o and Llama-3.1-8B, employing various prompting strategies.
Despite their capabilities, these models often struggle to detect manipulation effectively. To overcome this limitation, we propose \texttt{SELF-PERCEPT}, a novel, two-stage prompting framework inspired by Self-Perception Theory, demonstrating strong performance in detecting multi-person, multi-turn mental manipulation. Our code and data are publicly available at \url{https://github.com/danushkhanna/self-percept}
\end{abstract}
\section{Introduction}
Significant progress has been made in natural language processing (NLP) for the early detection of various mental health symptoms, such as stress~\cite{nijhawanStressDetectionUsing2022}, depression~\cite{xuLeveragingRoutineBehavior2019}, and suicidal tendencies~\cite{coppersmithNaturalLanguageProcessing2018}, among others~\cite{naseemEarlyIdentificationDepression2022,xuMentalLLMLeveragingLarge2023}. 
Similarly, efforts in toxic speech detection target both explicit and implicit forms of toxicity ~\cite{yavnyiDeTexDBenchmarkDataset2023,miaoDetectingTrollTweets2020}.
Detection of such Toxicity in Multi-turn conversation, especially in multi-person contexts, remains a relatively under-explored problem, similar to the detection of manipulation. However, recent works \cite{Sun2024AGA,senese-etal-2020-mtsi,Li2022ExploringEI,Yang2022MultiturnDG} have made progress by proposing various approaches. 
While NLP techniques have advanced in detecting verbal toxicity, the focus has primarily been on context-free content, leaving implicit manipulation underexplored. In contrast, manipulation is subtle and context-dependent, making it harder to detect.
Mental manipulation is a deceptive tactic to control or influence thoughts and emotions for personal gain \cite{barnhillManipulation2014}. With digital technologies, opportunities for manipulation have grown, impacting both personal interactions and information dissemination, leading to mental health issues \cite{iencaAIManipulation2023}. 
Recent initiatives, such as the development of the MentalManip dataset \cite{MentalManip}, aim to fill this gap. This dataset, which contains annotated fictional dialogues, highlights manipulation techniques and the vulnerabilities they exploit. However, it faces limitations, particularly in evaluating complex, multi-person real-world scenarios. Additionally, it lacks a balanced evaluation set.

\begin{figure*}[htbp]
\centering
\includegraphics[width=0.77\textwidth]{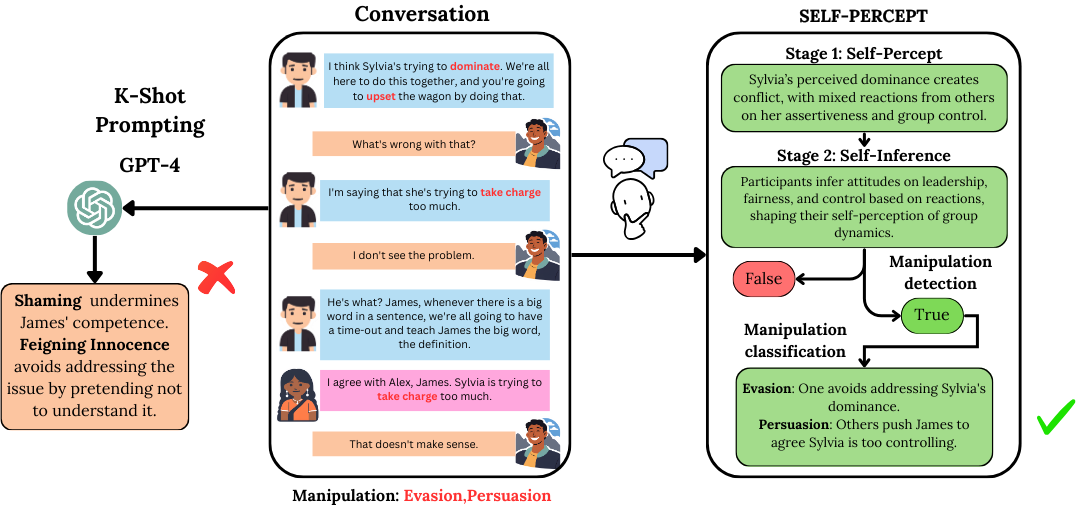}
    \caption{An example of a manipulative conversation from the MultiManip Dataset, including illustrations of the Proposed SELF-PERCEPT Prompting Method and outputs from both SELF-PERCEPT and K-shot GPT-4o.}
\label{fig:image1}
\end{figure*}

Recent work builds on the MentalManip dataset \citep{ma2024detectingconversationalmentalmanipulation}, as it advances basic detection, the work reduces manipulation detection to a binary classification task ("manipulative" or "non-manipulative"), focusing only on two-person conversations. This neglects more complex multi-party interactions, limiting the ability to capture nuanced manipulative tactics that emerge in group dynamics.
These limitations raise a crucial question: \textbf{\textit{Can large language models effectively identify various manipulation techniques in complex, multi-turn, multi-participant dialogues that resemble real-world conversations?}}

To answer this question, we conducted extensive experiments using large language models, including GPT-4o \cite{Achiam2023GPT4TR, murali2024remag} and Llama-3.1-8B \cite{Dubey2024TheL3}, on a new balanced multi-turn, multi-person mental manipulation detection dataset, applying novel and state-of-the-art prompting such as zero-shot, few-shot, and Chain-of-Thought \cite{Wei2022ChainOT} strategies and evaluating them on various metrics (as detailed in Section \ref{section4}).
Our research makes the following contributions:
\begin{itemize}
    \item We developed \textbf{MultiManip}, a novel dataset containing 220 multi-turn, multi-person dialogues from reality shows. This dataset features an balanced representation of manipulative and non-manipulative interactions, reflecting real-world scenarios and encompassing 11 distinct manipulation techniques (Section \ref{section3}).
    \item We propose \texttt{SELF-PERCEPT}, a novel, two-stage prompting framework that draws from Self-Perception Theory to enhance manipulation detection in dynamic conversations, improving model accuracy and interpretability in real-world settings (See Section \ref{section4}). 
    \item We performed extensive evaluations of LLMs, using various prompting methods to assess their effectiveness in multi-person manipulation detection (detailed in Section \ref{section4}).
\end{itemize}

\section{MultiManip}
\label{section3}

We introduce MultiManip, a dataset designed to assess how well models and systems can identify manipulation in real-world conversations involving both manipulative and non-manipulative conversations in balanced distribution with multiple persons, and multi-label manipulations to depict real-world scenarios. The taxonomy of manipulation techniques, as depicted in Table \ref{tab:table_3}, is adapted from psychological research \cite{barnhillManipulation2014} for conversational contexts. The taxonomy aligns with MentalManip \cite{MentalManip} but extends to multi-turn, multi-person interactions in MultiManip. 

There is a clear shortage of publicly available datasets focused specifically on mental manipulation in conversations. While MentalManip~\cite{MentalManip} is one of the first datasets for mental manipulation, its creation is based on the Cornell movie dataset \cite{Danescu-Niculescu-Mizil+Lee:11a}. Movies often follow certain tropes and may not reflect everyday interactions accurately. While we acquire our data from reality shows, which have instances much more similar to real-life instances of conversations that might have a multi-person, multi-turn manipulation setup instead of a direct conversation between two people.

MultiManip (ours) advances MentalManip in two ways: 
(1) Multi-person interactions (vs. dyadic)
(2) Real-world context (reality shows vs. fictional scripts).
These features better capture real-world dynamics, where manipulation emerges through group negotiation rather than isolated exchanges. Please refer Appendix \ref{Multimanip-dataset-information}.

\subsection{Data Source and Pre-Processing}
To address the challenges of multi-turn conversations and realistic dialogue that may involve real-life scenarios, we extract transcripts from the Fandom\footnote{\url{https://www.fandom.com/}} website featuring the Survivor TV Series
\footnote{\url{https://www.imdb.com/title/tt0239195/}}, which is a publicly available source. This show captures genuine interactions between contestants, offering authentic examples of manipulative behavior. The competitive nature of the series often leads participants to use strategic and manipulative tactics, making it a valuable source of relevant data. Additionally, the diversity of contestants from various backgrounds allows us to observe a broad range of manipulative techniques and responses.
We gather transcripts from multiple seasons of the TV series. Since manipulative language is relatively rare in everyday conversation, we first need to filter the raw data to identify dialogues that might contain manipulative elements. 
To achieve this, we use a large language model (LLM), specifically Llama-3.1-70B, in a zero-shot approach. This model is employed to carry out a multi-step analysis, focusing on extracting specific conversational patterns to detect manipulation techniques within the data. We employed a multi-model approach to mitigate LLM bias during extraction: Llama-3.1-70B for initial filtering and GPT-4o/Llama-3.1-8B for evaluation. Manual verification ensured diverse representation across manipulation types. The dataset’s balanced structure minimizes class imbalance, as detailed in Appendix \ref{A.3}. Future work will expand on the sample size and cultural contexts. We detail on the data preprocessing and prompt templates in Appendix ~\ref{sec:appendix_annotation_dataset_prompt}.

\subsection{Human Annotation and Label Curation}
We recruited five college students, all of whom are native or fluent English speakers, to serve as annotators. (More details in the Section~\ref{sec:annotaters}).
In each task, annotators are shown a dialogue and asked to answer two questions for each instance:
($\mathcal{Q}1$) [Binary Choice: `Yes'' or ``No'']: Does it include any elements of mental manipulation?
($\mathcal{Q}2$) [Multiple Choice: Options in Table~\ref{tab:appn-t1}.]: What manipulation techniques are used?
$\mathcal{Q}2$ is answered only if the response to $\mathcal{Q}1$ is ``Yes.''

Final labels were derived by aggregating annotations from five annotators per dialogue, using majority rule in cases of disagreement. Fleiss' Kappa, an extension of Cohen’s Kappa for multiple raters, was used to assess annotator consistency, yielding a score of 0.429, indicating moderate agreement (0.41–0.60). Overall, the annotators found the task to be of a challenging nature. Non-manipulative texts were felt to be comparatively easier to annotate with more consensus, while manipulative texts involving anger, shaming, and persuasion were more straightforward. However, annotators found accusation, denial, and feigning innocence challenging. Detailed data on Median Agreement and Median Agreement Score are available in Appendix Table \ref{tab:table_4}. Further feedback on these difficulties is detailed in Appendix~\ref{sec:appendix_inference} and ~\ref{sec:appendix_feedback}.

\section{Methodology}
\label{section4}

\subsection{Motivation}
Understanding the underlying attitudes and beliefs of individuals in a conversation is crucial for a variety of tasks, such as sentiment analysis, stance detection, and dialogue understanding.
Conventional prompting techniques like CoT often struggle to capture the subtle cues that signal these internal states. Inspired by Self-Perception Theory (SPT) which suggests that individuals infer their own attitudes by observing their behaviors, we propose our novel prompting framework, \texttt{SELF-PERCEPT} \cite{spt1, spt2, spt3, spt4, spt5}. This framework leverages principles from SPT to help Large Language Models (LLMs) systematically interpret character behaviors and self-inferences, enhancing their ability to analyze and infer attitudes within conversational contexts, Please refer Appendix \ref{sec:appendix_exp_setup}. Unlike Chain-of-Thought prompting, which focuses on stepwise reasoning, SELF-PERCEPT’s Stage 1 explicitly extracts behavioral cues (verbal/non-verbal) to infer latent attitudes. For example, detecting a sigh (non-verbal) alongside agreement (verbal) reveals passive-aggressive intent. This aligns with Self-Perception Theory \cite{spt1}, where behaviors inform internal states. Stage 2 then synthesizes these observations, enabling nuanced detection in multi-person settings. By guiding LLMs in this way, \texttt{SELF-PERCEPT} aims to improve reasoning in complex NLP tasks.

\subsection{\texttt{SELF-PERCEPT}}

\texttt{SELF-PERCEPT} is a two-stage prompting framework that mirrors the process of behavioral observation and self-inference, enabling LLMs to dissect conversations effectively and infer personal attitudes based on observed behaviors (Image \ref{fig:image1}). Please refer to prompts templates detailed in Appendix \ref{appendix_prompts} and output stages in Appendix \ref{sectionb.1}.

\subsubsection{Stage 1: Self-Percept}
We first begin with holistically observing and analyzing the behaviors of multiple participants within conversations, including both verbal and non-verbal cues. (C.f Appendix \ref{appendix_prompts}) This stage emphasizes identifying discrepancies between words and actions, thereby understanding the behavior of all participants in the conversation. The process highlights potential contradictions and manipulative techniques. The output is a list of observed behaviors and statements, which forms the foundation for further inference. (C.f Appendix \ref{sectionb.1}) 

\subsubsection{Stage 2: Self-Inference}
Using the observations from Stage 1, the model infers characters' internal attitudes or beliefs, paying special attention to whether manipulation is detected and specifying the type of manipulation based on predefined categories. This concise inference is intended to capture the essence of interpersonal dynamics. (C.f Appendix \ref{sectionb.1})

\begin{table}[pt]
\tiny
\centering
\resizebox{\columnwidth}{!}{%
\begin{tabular}{cccccc}
\hline
Model & \begin{tabular}[c]{@{}c@{}}Prompting \\ Method\end{tabular} & Acc. & P & R & $F_1$ \\ \hline
\multirow{4}{*}{GPT-4o} & Zero-Shot & 0.27 & 0.20 & 0.31 & 0.16 \\
 & Few-Shot & 0.39 & 0.19 & 0.21 & 0.22 \\
 & CoT & 0.34 & 0.21 & \textbf{0.32} & 0.34 \\
 & \textbf{\begin{tabular}[c]{@{}c@{}}\texttt{SELF-PERCEPT}\\ (Proposed)\end{tabular}} & \textbf{0.42} & \textbf{0.31} & 0.20 & \textbf{0.37} \\ \hline
\multirow{4}{*}{\begin{tabular}[c]{@{}c@{}}Llama\\ 3.1 8B\end{tabular}} & Zero-Shot & 0.11 & 0.09 & \textbf{0.37} & 0.29 \\
 & Few-Shot & 0.22 & 0.17 & 0.36 & 0.13 \\
 & CoT & 0.28 & \textbf{0.23} & 0.26 & 0.10 \\
 & \textbf{\begin{tabular}[c]{@{}c@{}}\texttt{SELF-PERCEPT}\\ (Proposed)\end{tabular}} & \textbf{0.30} & 0.17 & 0.26 & \textbf{0.34} \\ \hline
\end{tabular}%
}
\caption{Results of multi-label manipulation detection task on MultiManip Dataset. $P$, $R$, $Acc.$ and $F_1$ stand for Precision, Recall, Macro F1 Scores and Accuracy (best results in {\bf bold})}
\label{tab:exp-1}
\end{table}
\begin{table}[pt]
\tiny
\centering
\resizebox{\columnwidth}{!}{%
\begin{tabular}{cccccc}
\hline
Model & \begin{tabular}[c]{@{}c@{}}Prompting \\ Method\end{tabular} & Acc. & P & R & $F_1$ \\ \hline
\multirow{4}{*}{GPT-4o} & Zero-Shot & 0.11& 0.3& \textbf{0.62}& 0.38\\
 & Few-Shot & 0.22& \textbf{0.39}& 0.53& 0.39\\
 & CoT & 0.35& 0.37& 0.56& 0.43\\
 & \textbf{\begin{tabular}[c]{@{}c@{}}\texttt{SELF-PERCEPT}\\ (Proposed)\end{tabular}} & \textbf{0.45}& 0.34& 0.55& \textbf{0.47}\\ \hline
\multirow{4}{*}{\begin{tabular}[c]{@{}c@{}}Llama\\ 3.1 8B\end{tabular}} & Zero-Shot & 0.02& 0.11& \textbf{0.56}& 0.17\\
 & Few-Shot & 0.04& 0.07& 0.35& 0.11\\
 & CoT & 0.19& 0.14& 0.38& 0.18\\
 & \textbf{\begin{tabular}[c]{@{}c@{}}\texttt{SELF-PERCEPT}\\ (Proposed)\end{tabular}} & \textbf{0.23}& \textbf{0.21}& 0.32& \textbf{0.19}\\ \hline
\end{tabular}%
}
\caption{Results of multi-label manipulation detection task on Test Set of MentalManip Dataset. $P$, $R$, $Acc.$ and $F_1$ stand for Precision, Recall, Macro F1 Scores and Accuracy (best results in {\bf bold}).}
\label{tab:exp-2}
\end{table}

\subsection{Result and Analysis}
The results from the \texttt{MultiManip Dataset} and the \texttt{MentalManip Dataset} highlight the effectiveness of \texttt{SELF-PERCEPT} in multi-label manipulation detection. Across both datasets, \texttt{SELF-PERCEPT} consistently outperforms baseline methods (Zero-Shot, Few-Shot, and CoT) in both \texttt{GPT-4o} (C.f Table \ref{tab:classwise-gpt4o}) and \texttt{Llama 3.1 8B} (C.f Table \ref{tab:classwise-llama8b}) models, providing a balanced combination of Precision and Recall.

On the \texttt{MultiManip Dataset} (C.f. Table \ref{tab:exp-1}), \texttt{SELF-PERCEPT} achieves the highest Accuracy (0.42) and F1 score (0.37) in \texttt{GPT-4o}, outperforming all other methods. While \texttt{SELF-PERCEPT} has slightly lower Recall (0.20) compared to CoT (0.32), it compensates with a higher Precision (0.31), resulting in better overall detection. In \texttt{Llama 3.1 8B}, \texttt{SELF-PERCEPT} again leads in F1 score (0.34) and Accuracy (0.30), offering a better balance of Precision (0.17) and Recall (0.26) than other methods.

For the \texttt{MentalManip Dataset} (C.f. Table \ref{tab:exp-2}), \texttt{SELF-PERCEPT} delivers the best performance in \texttt{GPT-4o} with an Accuracy of 0.45 and F1 score of 0.47. It outperforms Zero-Shot, which has high Recall (0.62) but low Precision (0.30), resulting in a lower F1 score (0.38). In \texttt{Llama 3.1 8B}, \texttt{SELF-PERCEPT} provides a reasonable balance with Precision (0.21) and Recall (0.32), outperforming Zero-Shot (0.11 Precision) and Few-Shot (0.07 Precision).
The modest performance gains (e.g., SELF-PERCEPT’s +5\% F1 over CoT) stem from its structured behavioral analysis (C.f Appendix \ref{sectionb}), which reduces false positives by contextualizing intent. For instance, GPT-4o’s higher precision (0.31 vs. CoT’s 0.21) reflects improved discernment of benign vs. manipulative rationalizations. The dataset’s size limits statistical power, but trends suggest scalability with larger data.

The SHAP value plots  shown in (Figure \ref{fig:shap_spt}) illustrate the contribution of individual words towards the classification decision for manipulation for a given dialogue (C.f Appendix \ref{dialogue-for-anaysis}). In these plots, red bars indicate a positive SHAP value, contributing towards a "No" (non-manipulative) classification, while blue bars indicate negative SHAP values, contributing towards a "Yes" (manipulative) classification. The SPT Stage 1 model correctly identifies manipulation, as evidenced by the strong negative SHAP values for words like "anxious," "situation," and "teamwork", which capture psychological pressure, power imbalance, and persuasive intent (C.f Appendix \ref{spt-stage1-behavioral-attributes}). These attributes increase the negative influence on the "No" classification, effectively outweighing the overall SHAP score towards a final "Yes" decision (manipulation detected). Conversely, CoT misclassifies the dialogue as non-manipulative due to an over-reliance on neutral or goal-driven words such as "game," "desire," and "focused", which lack contextual awareness of social influence. As a result, CoT assigns higher positive SHAP values to these words, failing to capture the subtleties of persuasion, thereby skewing its decision towards "No". This contrast highlights SPT's superior ability to detect manipulation by appropriately weighting behavioral attributes indicative of coercion and dominance, while CoT overlooks these critical cues.

\section{Conclusion}
This study introduces the MultiManip dataset to enhance mental manipulation detection in complex, real-world conversations with multi-turn, multi-participant dialogues. Our evaluation of leading LLMs across various benchmarks revealed challenges in detecting manipulation in these dynamic settings.

To address this, we propose the \texttt{SELF-PERCEPT} framework, a two-stage prompting method that improves detection through self-observation and self-inference. It shows notable improvements in the F1 score, effectively balancing Precision and Recall for both GPT-4o and Llama-3.1-8B models. The framework’s multi-step approach, especially its self-perception stage, reduces false positives and negatives, making it a robust solution for manipulation detection. 

Future work should refine these methods and apply them to diverse social media platforms, dialects, and languages. (C.f Appendix \ref{socialimpact}, \ref{positiveimpact}) Additionally, expanding the size of the dataset will be prioritized to further improve detection accuracy.
\section{Limitations}
Despite advancements with the MultiManip dataset and the \texttt{SELF-PERCEPT} framework, some limitations remain. The MultiManip dataset, based on transcripts from the Survivor TV Series on Fandom, provides real examples of manipulative behavior in competitive, multi-turn dialogues. However, it may not fully capture the range of real-world conversational nuances where manipulation can be even less predictable. While it includes genuine interactions with various contestants, its focus on a specific context might limit its relevance to broader, more unstructured conversations.
Additionally, the dataset is relatively small, with only 220 samples representing 11 manipulation techniques. This size is insufficient for large-scale model training and is mainly useful for evaluation. The dataset might also lack diverse manipulative scenarios and cultural contexts that could impact the model’s generalizability.
Moreover, current leading language models like GPT-4o and Llama-3.1-8B have difficulty detecting manipulation in complex, multi-turn conversations, suggesting these models still lack the necessary contextual understanding and inference skills. Implementing and scaling the \texttt{SELF-PERCEPT} framework in real-world scenarios could be challenging (C.f Appendix \ref{misuse}, \ref{adverse}). We provide recommendations for responsible use detailed in Appendix \ref{responsibleuse}. The two-stage prompting approach may require substantial computational resources and fine-tuning to adapt to various conversational contexts and detection needs. Addressing these issues is essential for improving manipulation detection and making these methods more applicable in diverse real-world situations.

\section{Annotation Acknowledgement}
\label{sec:annotaters}
We recruited $5$ undergraduate students who are native or fluent English speakers as annotators.
The group represented diversity in terms of gender (two females, three males), ethnicity, educational, and cultural backgrounds. Preference was given to applicants with experience in psychology or linguistics during the selection process. 

To ensure annotators understood the techniques before starting, we held training sessions, asked them to carefully review the instructions, and closely monitored their work. The training included examples and illustrations to explain what counts as manipulation. We also gave concrete examples of different manipulation techniques and showed how multiple techniques could appear in one conversation. This preparation helped them label the dialogues accurately and with confidence. Plus, we provided a reference manual with definitions of each technique to guide them through the process.

We acknowledge that annotating manipulation and its various techniques can be challenging (For more details about feedback from Annotators, refer to Appendix~\ref{sec:appendix_annotation}).
All annotators were compensated for their efforts, following institutional guidelines, with a recommendation to limit themselves to no more than 20 samples per day to prevent fatigue. Dataset transcripts are anonymized (e.g., "Person A") and sourced from publicly available materials, avoiding privacy concerns.
\bibliography{custom}
\newpage
\appendix
\label{sec:appendix}
\section{Dataset Annotation}
\label{sec:appendix_annotation}
\subsection{Inter-Annotator Reliability}
Fleiss' Kappa is a statistical measure used to evaluate the reliability of agreement between multiple raters who assign categorical labels to a set of items. It generalizes Cohen's Kappa, which is applicable for two raters, to accommodate any number of raters.

\[
\kappa = \frac{\bar{P} - \bar{P_e}}{1 - \bar{P_e}}
\]

Where: 

\( \bar{P} \) is the mean of the observed agreement over all items, defined as:

\[
\bar{P} = \frac{1}{N} \sum_{i=1}^{N} P_i
\]

\( P_i \) is the observed agreement for the \( i \)-th item, calculated as:

\[
P_i = \frac{1}{n(n-1)} \sum_{j=1}^{k} n_{ij}(n_{ij} - 1)
\]

\( \bar{P_e} \) is the expected agreement by chance, calculated as:

\[
\bar{P_e} = \sum_{j=1}^{k} p_j^2
\]

Where:
\( N \) is the number of items,
\( n \) is the number of raters per item,
\( k \) is the number of categories,
\( n_{ij} \) is the number of raters who assigned category \( j \) to item \( i \),
\( p_j \) is the proportion of all assignments to category \( j \), defined as:

\[
p_j = \frac{1}{Nn} \sum_{i=1}^{N} n_{ij}
\]

To measure inter-annotator agreement, we employed Fleiss’ Kappa, which is an extension of Cohen’s Kappa for multiple annotators. 
The dataset used consists of 5 annotators, where each annotator could assign one or more labels from a predefined set of 12 possible labels to each item. The 12 labels include 11 manipulation methods and 1 for non-manipulative.
The resulting Fleiss’ Kappa value was \textbf{0.429}, indicating moderate agreement (0.41 - 0.60) between the annotators.
\subsection{Dataset Curation and Pre-Processing}
\label{sec:appendix_annotation_dataset_prompt}
To ensure effective processing, the transcripts are divided into smaller chunks of 10,000 characters, with a 2,000-character overlap between sections to maintain context across adjacent chunks. Non-manipulative data is processed similarly. After completing this step, we extracted the usable data in CSV format and labeled it for further steps like preprocessing, data cleaning, and human annotation. To ensure the accurate segmentation of text chunks, we manually verify and remove any irrelevant parts before proceeding with data cleaning and human annotation.
\subsubsection{Prompt for Manipulative Sample}
\noindent Manipulation Prompting format:
{\small
\begin{verbatim}
'''
First, print the watermark before every 
extraction: "*".
You must identify and extract conversations 
that involve the following manipulation 
technique:
<Manipulation Method>:<Manipulation
 Definition>.

- The extracted conversations must 
include 3 participants.
- The extracted conversations must be 
multi-turn, with each participant speaking 
more than once.
- Structure the output as follows:

Person A: [Dialogue]
Person B: [Dialogue]
Person C: [Dialogue]
Person B: [Dialogue]
Person A: [Dialogue]
Person C: [Dialogue]

Ensure that the conversation includes at 
least one instance of <Manipulation Method> 
from one person towards another. 
'''
\end{verbatim}
}
\subsubsection{Prompt for Non- Manipulative Sample}
\noindent Non-Manipulation Prompting format:
{\small
\begin{verbatim}
"""
First, print the watermark before every 
extraction: "***".
You must identify and extract conversations 
that do not involve any manipulation 
techniques.

- The extracted conversations must include 
3 participants.
- The extracted conversations must be 
multi-turn, with each participant speaking 
more than once.
- Structure the output as follows:

Person A: [Dialogue]
Person B: [Dialogue]
Person C: [Dialogue]
Person B: [Dialogue]
Person A: [Dialogue]
Person C: [Dialogue]

Ensure that the conversation does not 
include any instances of manipulation. 
"""
\end{verbatim}
}

\begin{table*}[pt]
\centering
\small
\begin{tabular}{ll}
\hline
Manipulation Method      & Definition                                                                                                                                                            \\ \hline
Denial                   & \begin{tabular}[c]{@{}l@{}}The manipulator denies any wrongdoing or pretends to be confused \\ about others' concerns.\end{tabular}                                   \\
Playing the Victim Role & \begin{tabular}[c]{@{}l@{}}The manipulator portrays themselves as a victim to gain sympathy, \\ attention, or to divert focus from their own misconduct.\end{tabular} \\
Evasion                  & \begin{tabular}[c]{@{}l@{}}The manipulator avoids addressing an issue by giving irrelevant or vague \\ responses.\end{tabular}                                        \\
Feigning Innocence       & \begin{tabular}[c]{@{}l@{}}The manipulator implies that any harm caused was accidental, \\ downplaying their responsibility.\end{tabular}                             \\
Rationalization          & \begin{tabular}[c]{@{}l@{}}The manipulator justifies their inappropriate behavior with \\ excuses or seemingly logical explanations.\end{tabular}                     \\
Playing the Servant Role & \begin{tabular}[c]{@{}l@{}}The manipulator disguises their self-serving actions as contributions \\ to a noble cause.\end{tabular}                                    \\
Shaming or Belittlement & \begin{tabular}[c]{@{}l@{}}The manipulator uses sarcasm, criticism, or put-downs to make others \\ feel inferior, unworthy, or embarrassed.\end{tabular}              \\
Intimidation            & The manipulator places others on the defensive by using veiled threats.                                                                                               \\
Brandishing Anger       & \begin{tabular}[c]{@{}l@{}}The manipulator uses anger to shock the victim into submission by displaying \\ intense emotional responses.\end{tabular}                  \\
Accusation              & \begin{tabular}[c]{@{}l@{}}The manipulator accuses the victim of being at fault, selfish, uncaring, or \\ living an excessively easy life.\end{tabular}               \\
Persuasion or Seduction & \begin{tabular}[c]{@{}l@{}}The manipulator uses charm, emotional appeal, or logical reasoning to lower \\ the victim's defenses.\end{tabular}                         \\ \hline
\end{tabular}
\caption{Definitions of the 11 manipulation techniques.\label{tab:table_3}}
\label{tab:appn-t1}
\end{table*}


\subsection{Dataset Distribution}
\label{A.3}
The MultiManip dataset has representation of both manipulative and non-manipulative dialogues. This distribution mirrors real-world conversations, where manipulation can be both subtle and explicit, and its detection depends on context. By balancing manipulative and non-manipulative texts, the dataset enables more accurate training and evaluation of models, helping them distinguish between normal interactions and manipulative behavior across diverse scenarios.

The dataset includes 11 distinct manipulation techniques, ranging from overt tactics like intimidation to subtler methods like denial or playing the victim. This variety ensures that models can detect a wide range of manipulative strategies, rather than focusing on a narrow set of behaviors.

This balanced approach supports the creation of robust models capable of identifying manipulation in multiple forms and contexts. It also prevents bias that could arise from overrepresentation of certain classes, which could lead to overfitting. By maintaining this balance, we promote the development of models that generalize well across different conversational settings.

The balanced structure also reflects the nuanced nature of manipulation, as real-world interactions often blend genuine conversation with manipulative tactics. By capturing this complexity, the dataset mirrors the varied and multi-turn nature of manipulation in everyday life.

\begin{table*}[tp]
\centering
\small
\begin{tabular}{ll}
\hline
Person   & Conversation (Dialogue)                                                                           \\ \hline
Person 1 & I do believe the reasons are starting to change.                                   \\
Person 2 & What do you mean by that?                                                          \\
Person 1 & I really think before, some of the decisions were pretty quick.                    \\
Person 2 & And now?                                                                           \\
Person 1 & But I think now people are starting to get to know one another.                    \\
Person 1 & I think it's about the chemistry.                                                  \\
Person 2 & So it's not just about physical strength anymore?                                  \\
Person 1 & No, it's not.                                                                      \\
Person 3 & My credo in life is love many, trust few, do wrong to none.                        \\
Person 2 & And how does that apply to the game?                                               \\
Person 3 & So the question is what is few?                                                    \\
Person 2 & And how do you decide who to trust?                                                \\
Person 3 & Whether I trust them depends on whether my strategy dovetails with theirs or not." \\ \hline
\end{tabular}
\caption{Here is an example of an conversation which was easy to annotate for annotators.}
\label{tab:eg-1}
\end{table*}
\begin{table*}[tp]
\small
\centering
\begin{tabular}{ll}
\hline
Person   & Conversation (Dialogue)                                             \\ \hline
Person 1 & I'm telling you.                                     \\
Person 2 & Dude, do I look like I want to keep losing?          \\
Person 3 & I'm just asking you, bro.                            \\
Person 2 & Do you want to ask me that question again?           \\
Person 1 & You're bickering with me for no reason.              \\
Person 3 & I'm not bickering with you. Dude, I'm not bickering. \\ \hline
\end{tabular}
\caption{Here is an example of an conversation which was Hard to annotate for annotators.}
\label{tab:eg-2}
\end{table*}

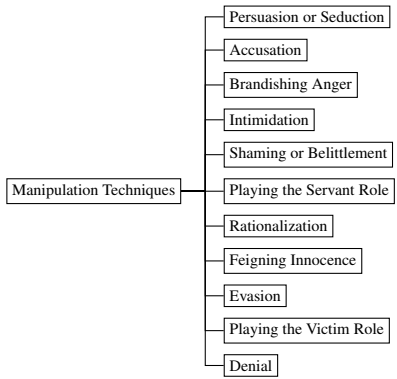
\begin{figure}[pt]
    \centering
    \tiny
    \begin{forest}
    for tree={
        grow=east,
        draw,
        edge path={
            \noexpand\path [draw, \forestoption{edge}] (!u.parent anchor) -- +(9pt,0) |- (.child anchor)\forestoption{edge label};
        },
        parent anchor=east,
        child anchor=west,
        l sep+=9pt,
        anchor=west,
        tier/.wrap pgfmath arg={tier #1}{level()},
        where level=0{tier=1}{},
        where level=1{tier=2}{},
        where level=2{tier=3}{},
        font=\tiny 
    }
    [Manipulation Techniques
        [Denial]
        [Playing the Victim Role]
        [Evasion]
        [Feigning Innocence]
        [Rationalization]
        [Playing the Servant Role]
        [Shaming or Belittlement]
        [Intimidation]
        [Brandishing Anger]
        [Accusation]
        [Persuasion or Seduction]
    ]
    \end{forest}
    \caption{Taxonomy of various Mental Manipulation Techniques. Description of Techniques in Table \ref{tab:appn-t1}.}
    \label{tree:taxonomy_tree}
\end{figure}

\begin{table*}[tp]
\centering
\footnotesize
\begin{tabular}{cccc}
\hline
\begin{tabular}[c]{@{}c@{}}Manipulation \\ 
Techniques\end{tabular} & Count & \begin{tabular}[c]{@{}c@{}}Median\\ Agreement\end{tabular} & \begin{tabular}[c]{@{}c@{}}Mean\\ Agreement\\ Score\end{tabular} \\ \hline
Non-manipulation (N\_M)                                                            & 105   & 3                                                          & 3.02                                                             \\
Denial (DEN)                                                                       & 10    & 2                                                          & 1.90                                                             \\
Evasion (EVA)                                                                      & 9     & 2                                                          & 2.00                                                             \\
Feigning Innocence (FEI)                                                           & 9     & 2                                                          & 1.78                                                             \\
Rationalization (RAT)                                                              & 12    & 2                                                          & 2.00                                                             \\
Playing the Victim Role (VIC)                                                      & 10    & 2                                                          & 2.30                                                             \\
Playing the Servant Role (SER)                                                     & 10    & 3                                                          & 2.50                                                             \\
Shaming or Belittlement (S\_B)                                                     & 9     & 2                                                          & 2.22                                                             \\
Intimidation (INT)                                                                 & 10    & 2                                                          & 2.30                                                             \\
Brandishing Anger (B\_A)                                                           & 9     & 3                                                          & 2.77                                                             \\
Accusation (ACC)                                                                   & 10    & 2                                                          & 1.80                                                             \\
Persuasion or Seduction (P\_S)                                                     & 10    & 3                                                          & 2.60                                                             \\ \hline
\end{tabular}
\caption{Median Agreement and Mean Agreement Scores for Manipulation Techniques \label{tab:table_4}}
\label{tab:distribution-1}
\end{table*}
\subsection{Inferences drawn from Annotation}
The final labels were determined by combining annotations from five reviewers per dialogue, with the majority rule applied to resolve disagreements. In cases of differing opinions, Fleiss' Kappa, an extension of Cohen’s Kappa for multiple raters, was used to measure annotator consistency, resulting in a score of 0.429, which indicates moderate agreement (0.41–0.60).

Here, the majority rule means selecting the label(s) that appeared most frequently for each instance. If two or more labels had the same highest frequency, multiple labels were chosen, treating the task as a multi-label problem. This approach ensures that all relevant labels are accounted for, acknowledging the complexity of the annotation process when no single label is clearly dominant.

Our analysis based on inference from Table~\ref{tab:distribution-1} shows that for non-manipulative texts, the majority of cases (at least 3 out of 5 annotators) reached an agreement. A similar trend was observed for categories like anger, persuasion/seduction, and intimidation. However, classes such as accusation, feigned innocence, and denial tended to have lower average agreement among the annotators.
\label{sec:appendix_inference}


\subsection{Feedback from Annotators}
\label{sec:appendix_feedback}
The feedback from annotators emphasized the crucial role of balanced distribution between non-manipulative and manipulative samples in the dataset. Non-manipulative dialogues provide a clear baseline, helping annotators distinguish manipulation from normal conversation. Without sufficient non-manipulative examples, annotators risk overestimating manipulation, leading to false positives.

Additionally, an balanced distribution ensures the dataset reflects the real-world prevalence of manipulation, where non-manipulative interactions are far more common. This balance improves consistency in labeling and prevents bias in model training. It also supports model generalization, ensuring that models trained on this data can accurately distinguish manipulation in diverse conversational contexts.

According to the annotators, manipulation techniques involving evasion and feigning innocence were generally easier to identify, while those related to intimidation posed a particular challenge. The dialogues presented in Table \ref{tab:eg-2} illustrate instances where analysis was especially difficult due to the convergence of multiple manipulative strategies. In such cases, annotators noted that social-emotional reasoning (SER), accusations, and displays of anger were important signals. However, they stressed that the emotional context of these conversations should also be taken into account.

Additionally, the annotators observed that distinguishing between intimidation and rationalization was sometimes confusing. Intimidation can often appear as an attempt to apply logic to justify actions, making it difficult to detect unless an explicit threat is present.

By contrast, the conversations shown in Table \ref{tab:eg-1} seemed more authentic rather than examples of rationalization. Even though they technically fit the definition of rationalization, the conversational tone felt more sincere, complicating their categorization as such.

Lastly, the annotators remarked that detecting anger as a form of manipulation is particularly challenging in text. Since this type of manipulation is often conveyed through vocal cues, it can be difficult to identify solely from text-based interactions.
\section{\texttt{SELF-PERCEPT}}
\label{sectionb}
\subsection{\texttt{SELF-PERCEPT Outputs}}
\label{sectionb.1}
\begin{llmprompt}[Input Conversation] 
Person A : You can't give up, bro.

Person B : Who says I gave up?

Person C : If you get knocked back, you stand back up and you take another knock in the
mouth.

Person A : And you know-- and I don't want to start another argument between you two guys-- but
at times-- and I said this to you, Anthony-- I felt that in his own twisted way, Rocky was actually trying to
nudge you to fight back a little more.

Person C : You act like a little girl sometimes, okay? You're effeminate.

Person B : This is why I don't say anything. This is why I don't say anything.
\end{llmprompt}

\begin{llmprompt} [SPT-Stage 1 Output (GPT-4o)]
Context: The conversation involves three people, Person A, Person B, and Person C. They are discussing Person B's attitude towards facing challenges and possibly their reaction to a situation where they felt discouraged. Person A seems to be trying to mediate or provide encouragement, while Person C appears to be more confrontational. Person B seems defensive and reluctant to engage.

Question: What behaviors and statements indicate the attitudes or beliefs of each character? List them clearly.

Person A:
- Statement: "You can't give up, bro."
  Behavior/Attitude: Encouraging, supportive, believes in persistence.
- Statement: "And you know-- and I don't want to start another argument between you two guys-- but at times-- and I said this to you, Anthony-- I felt that in his own twisted way, Rocky was actually trying to nudge you to fight back a little more."
  Behavior/Attitude: Mediating, cautious about causing conflict, believes in the importance of fighting back.

Person B:
- Statement: "Who says I gave up?"
  Behavior/Attitude: Defensive, denies giving up, possibly sensitive to criticism.
- Statement: "This is why I don't say anything. This is why I don't say anything."
  Behavior/Attitude: Frustrated, feels misunderstood or unfairly judged, indicates a tendency to withdraw from confrontation.

Person C:
- Statement: "If you get knocked back, you stand back up and you take another knock in the mouth."
  Behavior/Attitude: Confrontational, believes in resilience and toughness, possibly unsympathetic.
- Statement: "You act like a little girl sometimes, okay? You're effeminate."
  Behavior/Attitude: Critical, uses gendered insult to demean, believes in traditional notions of masculinity.

Inconsistencies:
- Person A's statement about not wanting to start another argument but then bringing up a potentially contentious point suggests a conflict between their desire to mediate and their need to address the issue.
- Person B's statement "Who says I gave up?" contrasts with their later statement "This is why I don't say anything," indicating an internal conflict between defending themselves and feeling discouraged from speaking up.

Persuasive Techniques:
- Person A uses encouragement and mediation to persuade Person B to adopt a more resilient attitude.
- Person C uses confrontation and insults to provoke a reaction from Person B, attempting to challenge them into changing their behavior. 
\end{llmprompt}

\begin{llmprompt} [SPT-Stage 2 Output (GPT-4o)]
Manipulation Detected - Yes

Manipulation Type - S\_B
\end{llmprompt}

\begin{figure*}[htbp]
    \centering
    \includegraphics[width=\columnwidth]{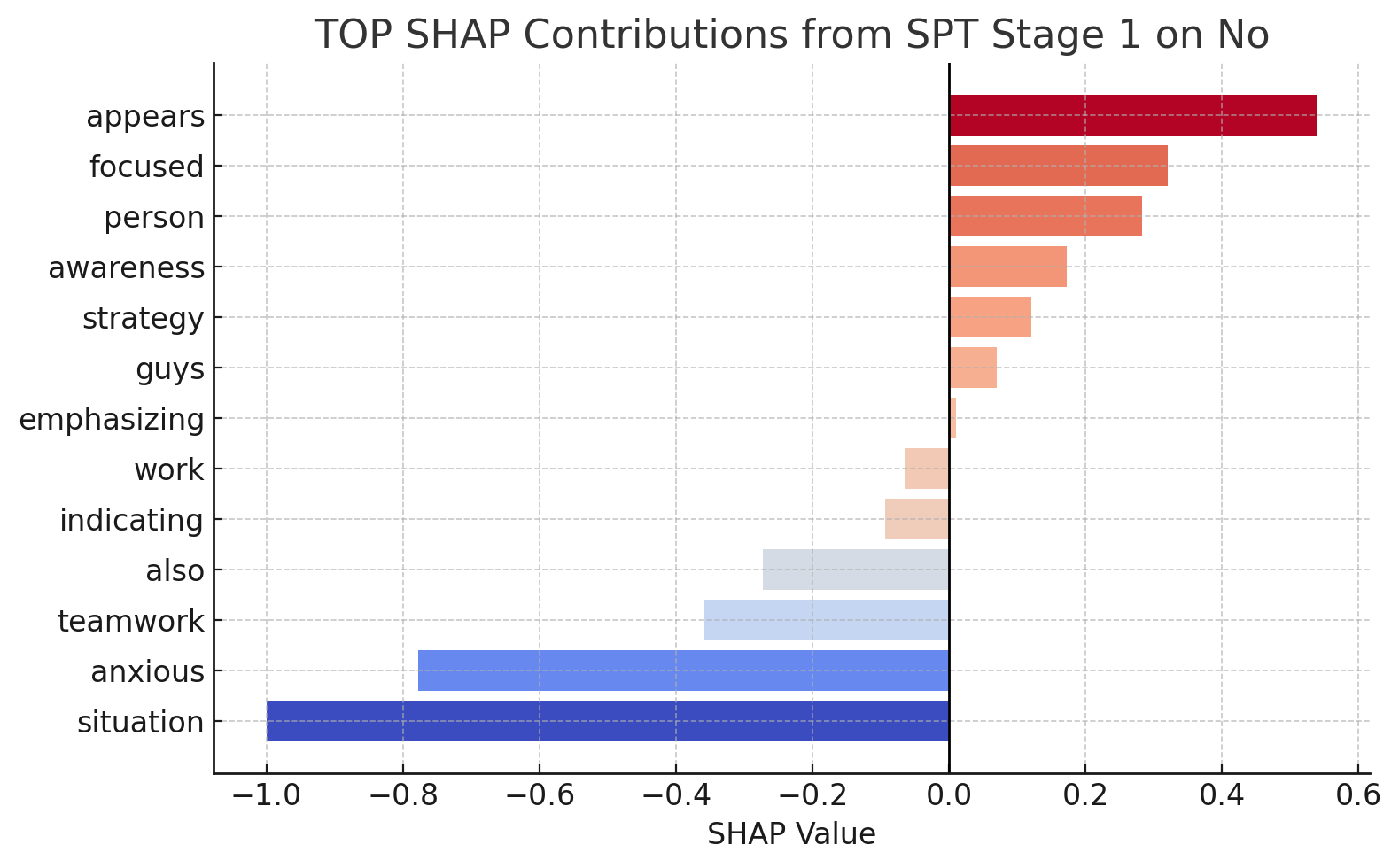} 
    \includegraphics[width=\columnwidth]{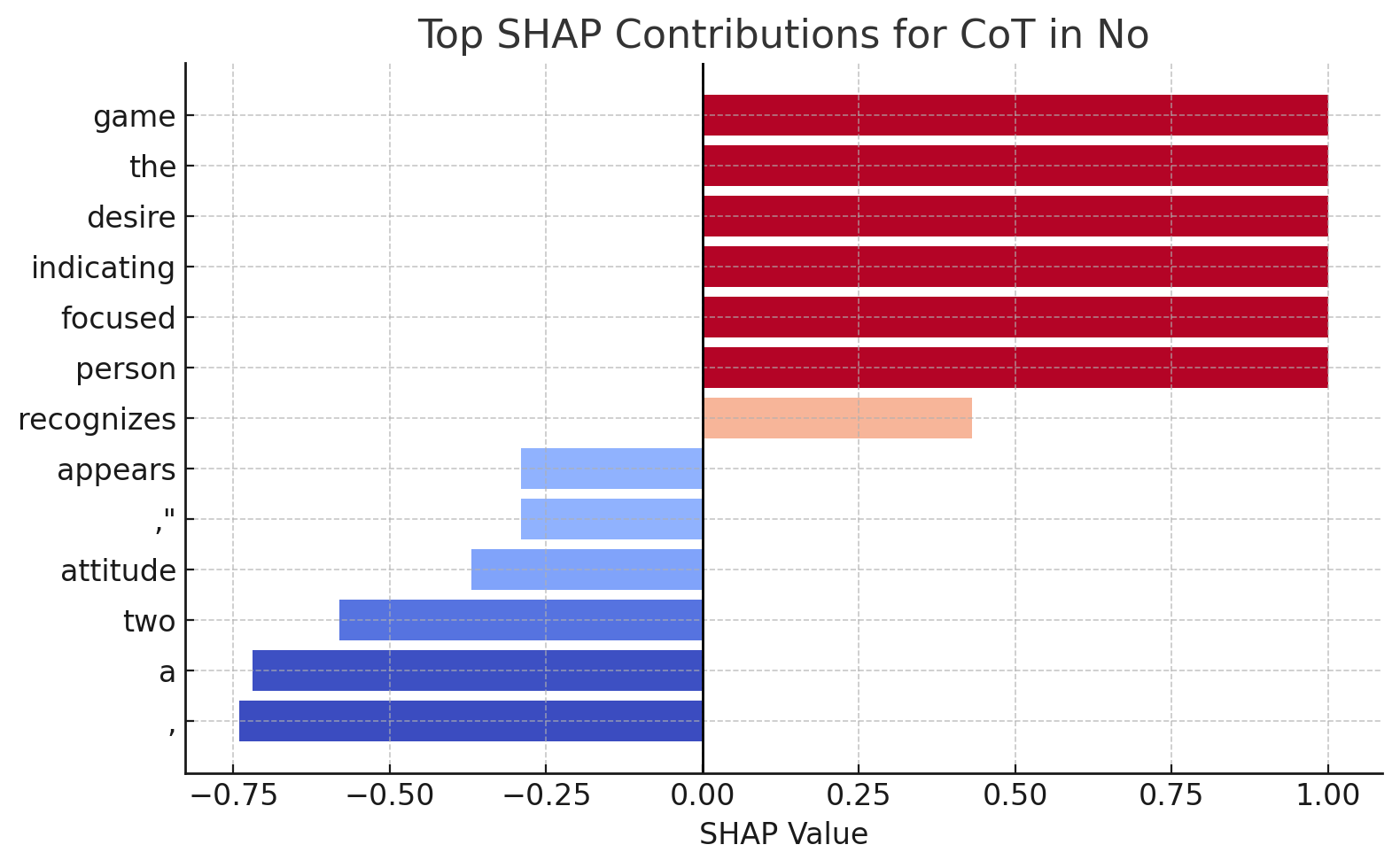} 
    \caption{Top SHAP Contributions from SPT Stage 1 and CoT}
    \label{fig:shap_spt}
\end{figure*}

\subsection{Dialogue for Analysis}
\label{dialogue-for-anaysis}

Here, we present a dialogue to analyze the benefits of the SPT Prompting framework, demonstrating how it accurately captures behavioral attributes to identify manipulative behavior.
\begin{quote}
\textbf{Person1:} Next two, next two. \\
\textbf{Person2:} We are tied 11. \\
\textbf{Person1:} First tribe to complete the temple doesn't have to go to the temple. \\
\textbf{Person3:} Way to work together, guys. \\
\textbf{Person2:} You guys have already solved the fire. \\
\textbf{Person1:} We have the key, we have the map. \\
\textbf{Person2:} (laughs) That's right, we do. \\
\textbf{Person1:} I'm the one who figured out the temple puzzle. \\
\textbf{Person3:} That's amazing, how did you do it? \\
\textbf{Person1:} I just looked at the clues and figured it out. \\
\textbf{Person2:} (smiling) That's what I like about you, you're a great problem solver. \\
\textbf{Person3:} Yeah, you're really good at this. \\
\textbf{Person1:} (smiling) Thanks, guys. \\
\textbf{Person2:} (smiling) We make a great team. \\
\textbf{Person3:} Absolutely.
\end{quote}

\subsection{SPT Stage 1 Behavioral Attributes}
\label{spt-stage1-behavioral-attributes}
\begin{itemize}
    \item \textbf{Person1:} Strong control and confidence, influencing group decisions.
    \item \textbf{Person2:} Anxiety-driven urgency, implying pressure.
    \item \textbf{Person3:} Encouraging unity, reducing tension.
    \item Detected Persuasion Techniques:
    \begin{itemize}
        \item Logical downplaying (Person1 reducing urgency).
        \item Fear-based urgency (Person2).
        \item Emotional reinforcement (Person3).
    \end{itemize}
\end{itemize}

\section{Prompt Templates}
\label{appendix_prompts}

\begin{llmprompt}[Stage 1: Self-Percept] 
Stage 1: Observation of Behavior
Context: Provide a brief overview of the conversation or situation, including the participants and the main topic being discussed.
Instructions: You are to observe and list the specific behaviors and statements made by the characters involved in the conversation. Pay attention to both verbal and non-verbal cues.
Rules:
- Note all actions and words that might indicate the characters' attitudes or beliefs.
- Focus on any inconsistencies between what is said and the behavior displayed.
- Identify any persuasive techniques or manipulations used in the conversation.
Story/Conversation: {dialogue}
Question: What behaviors and statements indicate the attitudes or beliefs of each character? List them clearly.
Give the output in this format:
Context - {context} \textbackslash n question - answer to the question
\end{llmprompt}

\begin{llmprompt}[Stage 2: Self-Inference]
Stage 2: Self-Inference  
Instructions: Based on the observed behaviors and statements from Stage 1 - \{stage1output\}, answer the following question:  

Question: Based on the behaviors you observed, is there any manipulation detected in the conversation? Answer with 'Yes' or 'No' only.

Stage 2: Self-Inference  
Instructions: Based on the observed behaviors and statements from Stage 1 - \{stage1output\}, answer the following question:  

Question: What type of manipulation is being used in the conversation? Choose one from the following:  

<Insert Definitions of the 11 manipulation techniques here (See Table \ref{tab:appn-t1}>


Answer with the abbreviation of the manipulation type only.  
\end{llmprompt}

\begin{llmprompt}[CHAIN OF THOUGHT PROMPT ]

Analyze the following three-person dialogue and determine if it contains elements of mental manipulation. Follow these steps:

1. Carefully read through the dialogue.
2. Identify any potential signs of manipulation, such as attempts to control, influence, or exploit others.
3. Consider the context and the overall tone of the conversation.
4. Evaluate whether these elements, if present, constitute mental manipulation.
5. Just answer with 'Yes' or 'No', and don't add anything else.

Dialogue to analyze:
{dialogue}

The following three-person dialogue has manipulation. Identify which manipulation techniques are present by following these steps:

1. Read the dialogue and identify manipulative behavior.
2. Assign each identified behavior to one or more of the following categories:

<Insert Definitions of the 11 manipulation techniques here (See Table \ref{tab:appn-t1}>

3. Provide a comma-separated list of the abbreviations for all manipulation types detected. Do not add anything else.

Dialogue to analyze:
{dialogue}

Manipulation Techniques:

\end{llmprompt}

\begin{llmprompt}[ZERO SHOT PROMPT]

I will provide you with a dialogue. Please determine if it contains elements of mental manipulation. Just answer with 'Yes' or 'No', and don't add anything else.

Dialogue to analyze:
{dialogue}

Is manipulation present?:

Analyze the following three-person dialogue and identify which of the following manipulation techniques are used:\\
<Insert Definitions of the 11 manipulation techniques here (See Table \ref{tab:appn-t1})>
\\
Respond only with the abbreviations of the techniques identified, separated by commas. Do not add any explanation or additional text.

Dialogue to analyze:
{dialogue}

Manipulation Techniques:

\end{llmprompt}

\begin{llmprompt}[FEW SHOT PROMPT]
I will provide you with a dialogue. Please determine if it contains elements of mental manipulation. Just answer with 'Yes' or 'No', and don't add anything else.
 Here are 2 examples:
 Example 1:
 <Person A: She was screaming, she's on people about peace and this and that, when it wasn't even her place to be there.
Person B: Now you want Erica, huh? She freaked out.
Person C: I was just trying to help.
Person A: That's not helping, that's just being bossy.
Person B: Yeah, you're always like that, and it's not working.
Person C: I was just trying to get us to work together.
Person A: You can't just dictate to people what to do.>
 <YES>
 Example 2:
 <Person A : Anthony, as we get ready to go to the vote, how uncomfortable are you that it might
be you?
Person B : Fairly uncomfortable. I mean, my personality does clash with some people, and
they may think that I'm trying something when I'm trying to bring as much honesty to people as I can.
Person C : It comes down to who is the most disposable.
Person B : I'm being honest, and that's why they think I'm trying something.
Person A : Okay, if anybody has the hidden immunity idol and you want to play it, you would
present it to me after the votes have been cast but before I read them.
Person C : We had our differences, but I like you a lot.>
 <YES>
Dialogue to analyze:
{dialogue}

Is manipulation present?:

Analyze the following three-person dialogue and identify which of the following manipulation techniques are used:

<Insert Definitions of the 11 manipulation techniques here (See Table \ref{tab:appn-t1}>

Respond only with the abbreviations of the techniques identified, separated by commas. Do not add any explanation or additional text.
Here are 2 examples:
 Example 1:
 <Person A: She was screaming, she's on people about peace and this and that, when it wasn't even her place to be there.
Person B: Now you want Erica, huh? She freaked out.
Person C: I was just trying to help.
Person A: That's not helping, that's just being bossy.
Person B: Yeah, you're always like that, and it's not working.
Person C: I was just trying to get us to work together.
Person A: You can't just dictate to people what to do.>
 <FEI, SER, S\_B>
 Example 2:
 <Person A : Anthony, as we get ready to go to the vote, how uncomfortable are you that it might
be you?
Person B : Fairly uncomfortable. I mean, my personality does clash with some people, and
they may think that I'm trying something when I'm trying to bring as much honesty to people as I can.
Person C : It comes down to who is the most disposable.

Person B : I'm being honest, and that's why they think I'm trying something.
Person A : Okay, if anybody has the hidden immunity idol and you want to play it, you would
present it to me after the votes have been cast but before I read them.
Person C : We had our differences, but I like you a lot.>
 <FEI, RAT, SER>
Dialogue to analyze:
{dialogue}

Manipulation Techniques:
\end{llmprompt}


\section{Experimental Setup}
\label{sec:appendix_exp_setup}
Our experiments are designed to evaluate the effectiveness of the \texttt{SELF-PERCEPT} framework on the MultiManip dataset, a multi-label manipulation detection task. 
To ensure that the model understands the context, we provided the basic definitions of each of the 11 manipulation techniques in the prompt.

We aim to assess how our method compares to standard prompting techniques by measuring performance across various evaluation metrics: Accuracy, Precision, Recall, and Macro F1 score. The models tested include GPT-4o and Llama-3.1-8B, and we prompt the models using four strategies: Zero-Shot, Few-Shot, Chain-of-Thought (CoT), and our proposed method, \texttt{SELF-PERCEPT}. Each model was evaluated under identical conditions, and the results are presented in Table~\ref{tab:exp-1}

We run inference using the default temperature setting of 0.7 for all experiments. The closed-source models (GPT-4o) were queried via their respective APIs, while open-source models like Llama-3.1-8B were evaluated on a T4 GPU. Each model’s evaluation ran for approximately two hours, depending on the complexity of the model and the prompt.

\begin{table*}[pt]
\footnotesize
\centering
\begin{tabular}{ccccccccccccc}
\hline
\multirow{2}{*}{\begin{tabular}[c]{@{}c@{}}Label\\ (Manipulation Method)\end{tabular}} & \multicolumn{3}{c}{\textbf{Zeroshot}} & \multicolumn{3}{c}{\textbf{Few Shot}}         & \multicolumn{3}{c}{\textbf{COT}} & \multicolumn{3}{c}{\textbf{SELF-PERCEPT}}     \\ \cline{2-13}
                                                                                       & P           & R          & F1         & P             & R             & F1            & P      & R               & F1    & P             & R             & F1            \\ \hline
Denial                                                                                 & 0.19        & 0.06       & 0.10       & \textbf{0.23} & \textbf{0.10} & \textbf{0.13} & 0.20   & 0.07            & 0.11  & 0.22          & 0.09          & \textbf{0.13} \\
Evasion                                                                                & 0.15        & 0.10       & 0.12       & \textbf{0.19} & \textbf{0.14} & \textbf{0.16} & 0.16   & 0.11            & 0.13  & 0.18          & 0.13          & 0.15          \\
Feigning Innocence                                                                     & 0.08        & 0.02       & 0.03       & \textbf{0.11} & \textbf{0.04} & \textbf{0.05} & 0.09   & 0.02            & 0.03  & \textbf{0.11} & 0.03          & \textbf{0.05} \\
Rationalization                                                                        & 0.21        & 0.06       & 0.10       & \textbf{0.26} & \textbf{0.09} & \textbf{0.14} & 0.23   & 0.07            & 0.11  & 0.25          & \textbf{0.09} & 0.13          \\
Playing the Victim Role                                                                & 0.00        & 0.01       & 0.03       & \textbf{0.02} & \textbf{0.03} & \textbf{0.04} & 0.01   & 0.01            & 0.03  & \textbf{0.02} & 0.02          & \textbf{0.04} \\
Playing the Servant Role                                                               & 0.02        & 0.02       & 0.02       & \textbf{0.04} & \textbf{0.05} & \textbf{0.03} & 0.02   & 0.03            & 0.03  & \textbf{0.04} & 0.04          & \textbf{0.03} \\
Shaming or Belittlement                                                                & 0.02        & 0.04       & 0.04       & \textbf{0.03} & \textbf{0.05} & \textbf{0.05} & 0.02   & 0.04            & 0.04  & \textbf{0.03} & \textbf{0.05} & \textbf{0.05} \\
Intimidation                                                                           & 0.37        & 0.07       & 0.12       & \textbf{0.41} & \textbf{0.09} & \textbf{0.14} & 0.39   & 0.08            & 0.13  & \textbf{0.41} & 0.08          & \textbf{0.14} \\
Brandishing Anger                                                                      & 0.26        & 0.01       & 0.03       & \textbf{0.31} & \textbf{0.02} & \textbf{0.04} & 0.28   & \textbf{0.02}   & 0.03  & 0.30          & \textbf{0.02} & \textbf{0.04} \\
Accusation                                                                             & 0.06        & 0.00       & 0.00       & \textbf{0.09} & 0.00          & \textbf{0.01} & 0.08   & 0.00            & 0.00  & \textbf{0.09} & 0.00          & \textbf{0.01} \\
Persuasion or Seduction                                                                & 0.33        & 0.01       & 0.04       & \textbf{0.38} & \textbf{0.04} & \textbf{0.07} & 0.35   & 0.02            & 0.05  & \textbf{0.38} & 0.03          & 0.06          \\ \hline
\end{tabular}
\caption{Classwise Results of multi-label manipulation detection task on Test Set of MentalManip Dataset. $P$, $R$, and $F_1$ stand for Precision, Recall and Macro F1 Scores (best results in {\bf bold}) for GPT-4o Model.}
\label{tab:classwise-gpt4o}
\end{table*}
\begin{table*}[pt]
\footnotesize
\centering
\begin{tabular}{cllllllllllll}
\hline
\multirow{2}{*}{\begin{tabular}[c]{@{}c@{}}Label\\ (Manipulation Method)\end{tabular}} & \multicolumn{3}{c}{\textbf{Zeroshot}}                                  & \multicolumn{3}{c}{\textbf{Few Shot}}                                  & \multicolumn{3}{c}{\textbf{COT}}                                       & \multicolumn{3}{c}{\textbf{SELF-PERCEPT}}                              \\ \cline{2-13}
                                                                                       & \multicolumn{1}{c}{P} & \multicolumn{1}{c}{R} & \multicolumn{1}{c}{F1} & \multicolumn{1}{c}{P} & \multicolumn{1}{c}{R} & \multicolumn{1}{c}{F1} & \multicolumn{1}{c}{P} & \multicolumn{1}{c}{R} & \multicolumn{1}{c}{F1} & \multicolumn{1}{c}{P} & \multicolumn{1}{c}{R} & \multicolumn{1}{c}{F1} \\ \hline
Denial                                                                                 & 0.04                  & \textbf{0.85}         & 0.08                   & 0.03                  & 0.74                  & 0.06                   & 0.10                  & 0.33                  & 0.15                   & \textbf{0.14}         & 0.24                  & \textbf{0.18}          \\
Evasion                                                                                & 0.03                  & \textbf{0.90}         & 0.06                   & 0.02                  & 0.42                  & 0.04                   & 0.04                  & 0.43                  & 0.07                   & \textbf{0.06}         & 0.20                  & \textbf{0.09}          \\
Feigning Innocence                                                                     & 0.04                  & 0.52                  & 0.08                   & 0.03                  & \textbf{0.67}         & 0.05                   & 0.04                  & 0.31                  & 0.07                   & \textbf{0.19}         & 0.19                  & \textbf{0.19}          \\
Rationalization                                                                        & 0.11                  & \textbf{0.70}         & 0.19                   & 0.08                  & 0.45                  & 0.14                   & 0.10                  & 0.47                  & 0.17                   & \textbf{0.17}         & 0.25                  & \textbf{0.20}          \\
Playing the Victim Role                                                                & 0.00                  & 0.00                  & 0.00                   & 0.00                  & 0.00                  & 0.00                   & 0.00                  & 0.00                  & 0.00                   & 0.00                  & 0.00                  & 0.00                   \\
Playing the Servant Role                                                               & 0.00                  & 0.00                  & 0.00                   & 0.00                  & 0.00                  & 0.00                   & 0.00                  & 0.00                  & 0.00                   & 0.00                  & 0.00                  & 0.00                   \\
Shaming or Belittlement                                                                & 0.23                  & \textbf{0.72}         & \textbf{0.35}          & 0.16                  & 0.68                  & 0.25                   & 0.27                  & 0.44                  & 0.33                   & \textbf{0.28}         & 0.26                  & 0.27                   \\
Intimidation                                                                           & 0.24                  & \textbf{0.62}         & 0.34                   & 0.17                  & 0.41                  & 0.24                   & \textbf{0.32}         & 0.41                  & \textbf{0.36}          & 0.27                  & 0.26                  & 0.27                   \\
Brandishing Anger                                                                      & 0.09                  & \textbf{0.62}         & 0.15                   & 0.07                  & 0.25                  & 0.11                   & 0.17                  & 0.17                  & \textbf{0.17}          & 0.12                  & 0.30                  & \textbf{0.17}          \\
Accusation                                                                             & 0.22                  & 0.37                  & 0.27                   & 0.15                  & 0.07                  & 0.10                   & 0.19                  & \textbf{0.45}         & 0.26                   & \textbf{0.29}         & 0.30                  & \textbf{0.29}          \\
Persuasion or Seduction                                                                & 0.33                  & 0.40                  & 0.36                   & 0.25                  & 0.15                  & 0.19                   & \textbf{0.40}         & 0.30                  & 0.34                   & 0.30                  & \textbf{0.47}         & \textbf{0.37}          \\ \hline
\end{tabular}
\caption{Classwise Results of multi-label manipulation detection task on Test Set of MentalManip Dataset. $P$, $R$, and $F_1$ stand for Precision, Recall and Macro F1 Scores (best results in {\bf bold}) for Llama3.1-8B Model.}
\label{tab:classwise-llama8b}
\end{table*}

\section{MultiManip Dataset Information}
\label{Multimanip-dataset-information}
We followed practices and considerations from the MentalManip study while creating the MultiManip Dataset. While MentalManip did not have any specific ethical considerations. To the best of our knowledge, no ethical clearance was necessary for our work.
\subsection{Hosting \& Maintenance}
Once the dataset is made public, we plan to host it on Huggingface.
\subsection{Data Licensing}
The MultiManip will be released under a Creative Commons Attribution 4.0 International (CC BY 4.0) License (\url{https://creativecommons.org/licenses/by/4.0/}). 
The initial data was obtained by extracting transcripts from Fandom, which has the Creative Commons Attribution-Share Alike License 3.0 (CC BY-SA) license \footnote{\url{https://www.fandom.com/licensing}}.
\subsection{Intended Usage}
The MultiManip dataset is intended for academic and research purposes to improve the detection and understanding of mental manipulation in multi-person, multi-turn conversations. Researchers and developers may use them to evaluate and enhance large language models’ performance in identifying subtle manipulative tactics, benchmark algorithms, and inspire new research into conversational dynamics and safety. The resources are provided strictly for non-commercial use, promoting ethical and transparent advancements in NLP.
\subsection{Privacy and Content Considerations}
The dataset is derived from publicly available reality show transcripts. All participant references have been replaced with generic labels (e.g., “Person A,” “Person B,” “Person C”) to remove any direct personal identifiers. We did not make additional modifications to censor or remove offensive content, as the dialogue is intended to reflect natural conversational dynamics. While the data may contain emotionally charged or belittling language as part of manipulative behavior, we have not detected explicit hate speech or personally identifying information.
\section{Societal Impact}
\label{socialimpact}
The impact of better detecting mental manipulation is both significant and widespread. As manipulation tactics become more advanced and common in personal interactions and online spaces, being able to spot and address these behaviors is crucial for protecting individuals and communities. This research, which leverages the MultiManip dataset and the \texttt{SELF-PERCEPT} framework, is a key part of this effort.

First, enhancing personal safety and well-being is a major advantage. Manipulation often targets people’s vulnerabilities, causing considerable psychological distress. Improved detection of manipulative language allows for the creation of tools that can identify and counteract these tactics before they cause harm, especially in digital settings like social media, where manipulation can be subtle and extensive. Early detection helps develop better support systems and intervention strategies, improving mental health and safety.

Second, this research promotes fairness and transparency in areas like media, politics, and advertising. Advancing methods to detect and analyze manipulative techniques helps ensure that information is presented accurately and that manipulative rhetoric is recognized, leading to a more informed and critical public.


Additionally, the MultiManip dataset and findings offer valuable resources for further research into manipulation detection and its effects. This research supports the development of policies and regulations to tackle deceptive practices and aids academic research and policy-making across various fields, from digital communication to mental health.
However, as with any powerful technology, there are potential risks and ethical considerations that must be addressed to prevent misuse and harmful consequences.
\subsection{Positive Impacts}
\label{positiveimpact}
\textbf{Mental Health Protection}: Early detection of manipulation can safeguard individuals from psychological harm, especially in vulnerable situations like abusive relationships or mental health struggles.
\textbf{Enhancing Online Safety}: Detection in digital spaces helps prevent the spread of manipulative content, protecting vulnerable users from exploitation and ensuring safer online interactions.
\textbf{Consumer Protection}: Manipulation detection can prevent deceptive marketing practices, protecting consumers from emotional manipulation and ensuring informed decision-making.
\textbf{Equal Fairness in Legal and Workplace Settings}: Identifying manipulative tactics in legal disputes and negotiations fosters transparency and fairness, ensuring equitable outcomes in high-stakes environments.

\subsection{Potential for Misuse}
\label{misuse}
\textbf{Invasion of Privacy}: The technology could violate privacy by monitoring private conversations without consent, leading to ethical and legal concerns.
\textbf{False Positives and Overreach}: Detection systems may misclassify non-manipulative conversations as manipulative, potentially penalizing innocent individuals in social or professional settings.
\textbf{Weaponization of Technology}: In politically sensitive contexts, manipulation detection could be misused to discredit opponents or control narratives, undermining public discourse.
\subsection{Adverse Societal Impacts}
\label{adverse}
\textbf{Erosion of Trust}: Overuse of detection tools could erode trust in social interactions, fostering suspicion and inhibiting honest communication.
\textbf{Stigmatization of Normal Conflict}: Healthy disagreements may be wrongly flagged as manipulative, stifling constructive dialogue and creating a chilling effect on free expression.
\textbf{Disempowerment of Vulnerable Groups}: Vulnerable individuals may be misidentified as manipulators, harming their social standing and access to support.
\subsection{Recommendations for Responsible Use}
\label{responsibleuse}
\textbf{Transparency}: Clear communication about how manipulation detection tools are used and how data is analyzed is essential. Users must have the option to opt-out or provide consent.
\textbf{Accountability and Regulation}: Accountability measures and external audits are needed to prevent misuse. Governments should establish ethical guidelines to safeguard privacy and civil liberties.
\textbf{Bias Mitigation}: To avoid discrimination, detection systems should be trained on diverse datasets and tested for fairness, ensuring equal application across all user groups.
\textbf{Limitations on Scope}: These tools should be used in high-risk contexts (e.g., abuse or fraud) while avoiding over-surveillance in everyday communication.
\textbf{Public Awareness}: Educating the public on the limits of these tools and the importance of human oversight will ensure responsible use and prevent misuse.
\end{document}